\documentclass{article}  
\usepackage[final]{corl_2020} 

\usepackage{amssymb}
\usepackage{graphicx,amsmath}
\usepackage{lineno}
\usepackage{caption,url}
\usepackage{float}
\usepackage{epstopdf}
\usepackage{authblk}

\usepackage[
  separate-uncertainty = true,
  multi-part-units = repeat
]{siunitx}

\usepackage{subcaption}
\usepackage{gensymb}

\def\@fnsymbol#1{\ensuremath{\ifcase#1\or *\or \dagger\or \ddagger\or
   \mathsection\or \mathparagraph\or \|\or **\or \dagger\dagger
   \or \ddagger\ddagger \else\@ctrerr\fi}}

\makeatletter
\newcommand{\ssymbol}[1]{^{\@fnsymbol{#1}}}
\newcommand{\R}{\mathbb{R}}
\makeatother

\begin{document}

\title{\LARGE \bf Robust Quadrupedal Locomotion on Sloped Terrains: A Linear Policy Approach
}



\author[]{Kartik Paigwar}
\author[]{Lokesh Krishna}
\author[]{Sashank Tirumala}
\author[]{Naman Khetan}
\author[]{Aditya Sagi}
\author[]{Ashish Joglekar}
\author[]{Shalabh Bhatnagar}
\author[]{Ashitava Ghosal}
\author[]{Bharadwaj Amrutur}
\author[]{Shishir Kolathaya}

\affil[]{\footnotesize Robert Bosch Centre for Cyber Physical Systems\\Indian Institute of Science, Bengaluru\\
stochlab@iisc.ac.in}


\maketitle
\thispagestyle{empty}
\pagestyle{empty}

\begin{abstract}
In this paper, with a view toward fast deployment of locomotion gaits in low-cost hardware, we use a linear policy for realizing end-foot trajectories in the quadruped robot, Stoch $2$. In particular, the parameters of the end-foot trajectories are shaped via a linear feedback policy that takes the torso orientation and the terrain slope as inputs. The corresponding desired joint angles are obtained via an inverse kinematics solver and tracked via a PID control law. Augmented Random Search, a model-free and a gradient-free learning algorithm is used to train this linear policy. Simulation results show that the resulting walking is robust to terrain slope variations and external pushes. This methodology is not only computationally light-weight but also uses minimal sensing and actuation capabilities in the robot, thereby justifying the approach.
\end{abstract}


\section{Introduction}

Over the last few years, there has been a surge in the development of low-cost and open-source quadruped robots \cite{Nathan2019Stanford}, \cite{grimminger2019open}, \cite{dhaivatdesigndevelopment}.
Due to limitations in actuation/sensing capabilities, developing a control law that yields stable walking in these types of robots is cumbersome. Some of these robots do not possess 
BLDC motors for accurate torque control \cite{dhaivatdesigndevelopment}. In addition, deployment of some of the advanced control laws like the model predictive control (MPC) \cite{Kim2019HighlyDQ}, Deep Neural Networks (DNN) \cite{google_paper,Hwangboeaau5872} require expensive computational resources.
Therefore, in this paper, we would like to answer the following question: What is the minimum possible control framework that can be deployed to realize stable locomotion behaviours in medium-size low-cost quadruped robots? 

The domain of quadrupedal locomotion has reached maturity today with quite a few research labs/companies successfully commercializing their quadrupeds. A slew of techniques are in use---inverted pendulum model-based controllers \cite{raibert1986legged}, zero-moment point- based controllers \cite{vukobratovic2004zero}, hybrid zero dynamics \cite{hzd_grizzle}, model predictive controllers \cite{Kim2019HighlyDQ}---to name a few.
However, these techniques are not suitable for our goal because they require extensive domain expertise.
The long term goal of our work is to create a quadruped controller akin to inexpensive flight controllers developed for remote piloting of autonomous aircrafts \cite{pixhawk2011,px42015}. The difficulty in developing such a controller is that it must be capable of adapting to different quadruped robots with minimum modifications. 

\begin{figure}[t!]
\centering
\includegraphics[height = 0.25\linewidth] {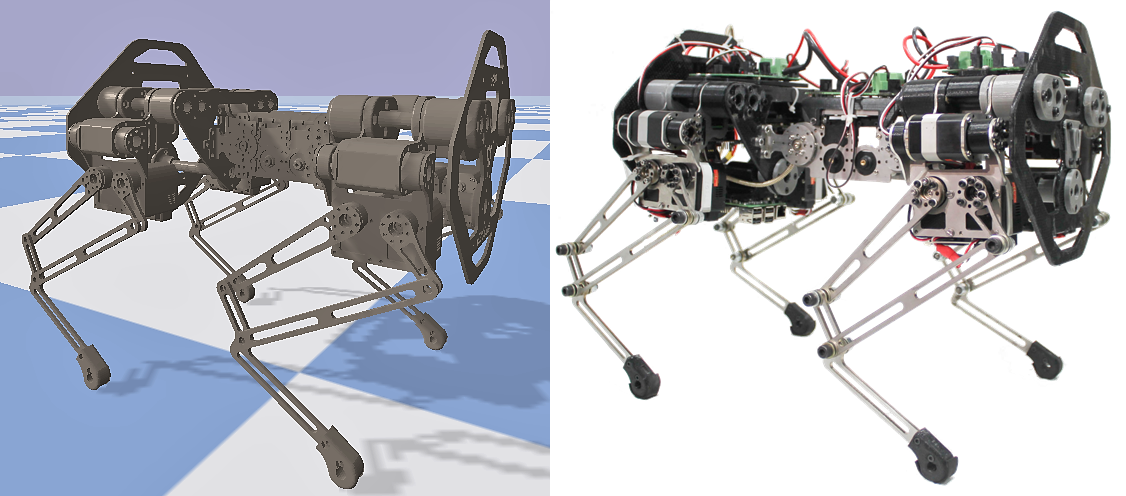}
\includegraphics[height = 0.25
    \linewidth]{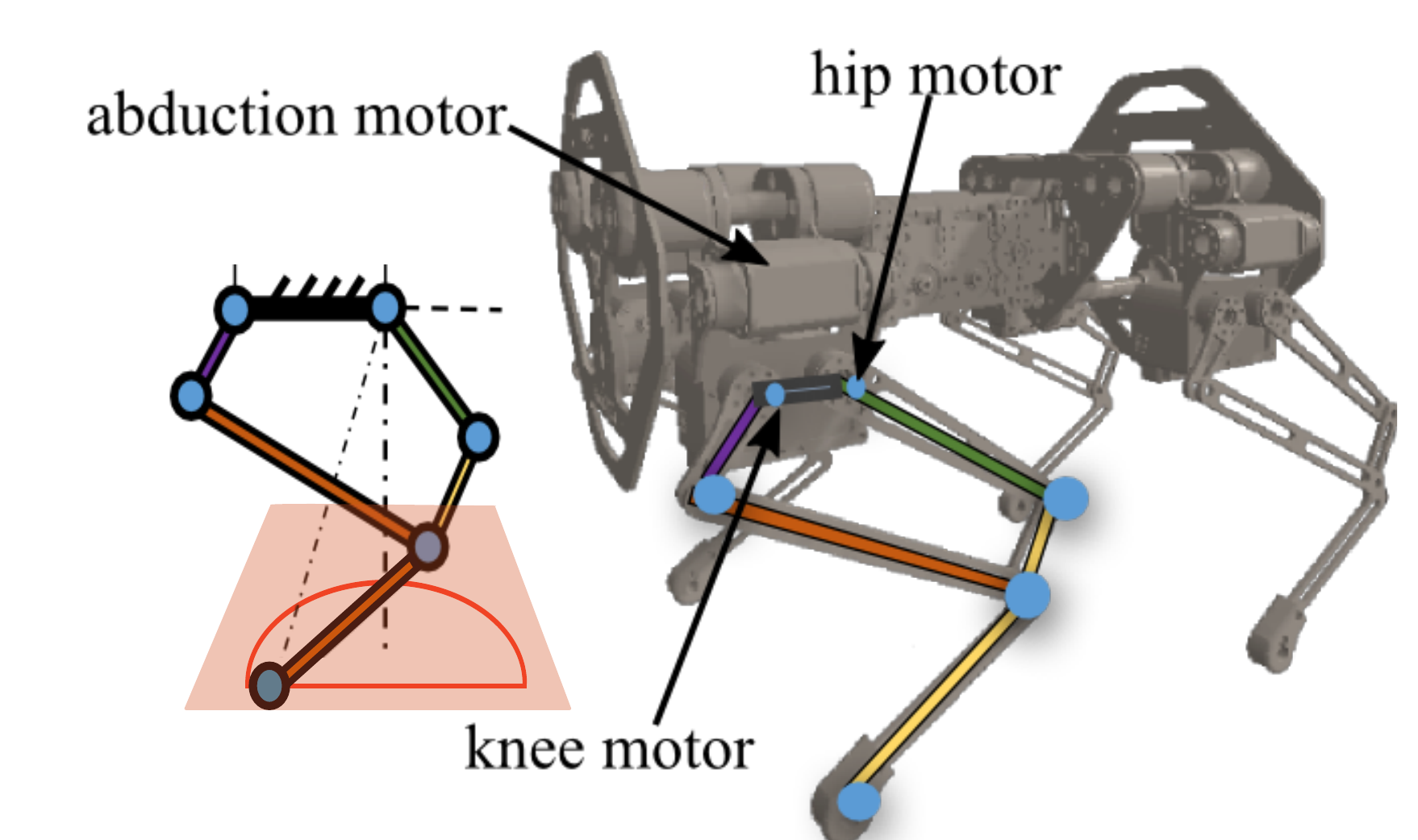}
    \caption{Simulated model and fabricated hardware of quadruped robot, Stoch 2, respectively (left two figures). The kinematic description of a leg (rightmost). 
    The trajectory followed by the foot is a  semi-ellipse which lies within a safe workspace of the leg which is shown as the trapezoidal region.}
\label{fig:IK_S2}
\vspace{-2mm}
\end{figure}

With a view toward ``automatically" developing control algorithms for a broad set of servo driven quadruped robots, we aim to learn a simple policy that outputs end-foot reference trajectories that can be tracked by the individual joints, resulting in stable walking.
To this end, data driven approaches like
Reinforcement Learning (RL)  \cite{kober2013reinforcement}, which are capable of learning walking controllers by themselves 
seem very attractive. 
In our formulation, we learn a linear policy, i.e. a single matrix which can be implemented on microcontroller based hardware boards (like the STM32 series or ARM Cortex M4 series) using RL and is capable of traversing slopes, rough terrain and rejecting disturbances. 

\subsection{Related work}

Reinforcement learning (RL) for quadrupedal walking was first explored by \cite{kohl2004policy}, where policy gradient algorithms were used to learn optimal parameters for end-foot trajectories. The learned trajectories were then tracked in each foot in open-loop.
A deep reinforcement learning framework (D-RL) was used in \cite{Hwangboeaau5872}, \cite{Jie2018Sim},  for training a deep neural network (DNN) based feedback policy for walking on flat terrains by first training the policy in simulation, and then bridging the sim-to-real gap with domain randomization and motor-modelling.
However, these DNN based policies are computationally intensive and were unable to 
generalize to rough terrain walking. Similarly \cite{PMTG}, \cite{Trajectory2019Kolathaya}, \cite{Sashank2020Gait}  trained a quadruped for a flat terrain, where the learning process was sped up significantly by parameterizing the control policy using central pattern generators, B\'ezier curves and rational B\'eziers respectively. In our paper, we also parameterize our control policy with elliptical curves to speed up training. \cite{rajeswaran2017towards} first demonstrated the capabilities of a linear policy for robotics control in simulation while \cite{tirumala2019gait} used the linear policy approach to create policies for flat terrain quadruped walking. We also use a similar linear policy paradigm for rough terrain quadruped walking.  
Walking on arbitrary terrains was formulated as a nonlinear optimization problem in \cite{winkler18}. \cite{melon2020reliable} learnt the parameter initialization for the above nonlinear optimization problem, improving the accuracy and speed of the optimization and enabling the quadruped robot to navigate bumpy terrain. We used a similar initialization technique in our training process. A custom RL algorithm was used in \cite{Tsounis2020Deep},  to enforce footstep constraints in order to train walking over stairs, slopes and ridges. Similarly \cite{Kolter2007Hierarchical} uses hierarchical apprenticeship learning and expert data to navigate a rocky terrain. 
However, these policies 
require a complete terrain map of as an input to work. Our focus is mainly in trotting on slopes 
with  proprioceptive feedback. 

Outside of learning, a large number of techniques have been developed to navigate a rough terrain for a quadruped robot in a blind manner. The MIT Cheetah robot achieved this by using Model Predictive Control for the stance legs and Raibert's controller for the swing legs \cite{Bledt2018MIT}. Similarly, \cite{Dario2017Dynamic} demonstrated a hierarchical blind whole body controller for the Anymal robot. However, these techniques require expensive Series Elastic Actuators or Quasi Direct-Drive actuators in order to function. In contrast, our approach requires only hobby servo motors and mainly focuses on reinforcement learning for rough terrain locomotion.

\subsection{Contributions and organization of the paper}
In this paper, we extend the work in \cite{PMTG}, \cite{Trajectory2019Kolathaya}, \cite{tirumala2019gait}   to include locomotion on rough terrain. In particular \cite{Trajectory2019Kolathaya} used open-loop trajectories that were tracked in each leg to yield walking. Our approach consist of a feedback control to dynamically shape these trajectories based on the body and plane orientation.
Our controller is capable of rejecting disturbances and maintaining balance on slopes of various inclines. 
The support plane is estimated using forward kinematics and foot contacts similar to \cite{Bledt2018MIT}. We also restrict our policy to being linear. This linearity requires low computation, thus enabling our policy to be run in real-time on an on-board embedded system. Augmented Random Search (ARS), a well-known algorithm for training linear policies \cite{mania2018simple}, is used to learn our proposed policy. We validate our results on the indigenously developed quadruped robot, Stoch $2$ in simulation. 




The paper is structured as follows: Section \ref{sec:background} will provide a description of the robot and the associated control framework, Section \ref{sec:learningalgorithm} will describe the training process used. This will be followed by simulation results in Section \ref{sec:results}.

\section{Robot model and problem formulation}
\label{sec:background}

In this section, we will discuss our custom built quadruped robot Stoch $2$. Specifically, we will provide details about the hardware, and the walking controller that we aim to learn with reinforcement learning.


\subsection{Hardware description of Stoch $2$}
\textit{Stoch $2$}  is  a second generation quadruped robot in the \textit{Stoch} series designed and developed at the Indian Institute of Science \cite{dhaivatdesigndevelopment}. It is a low cost hardware (costing less than \$$3000$) which is built using hobby servo motors and 3D-printed parts. It has an inertial measurement unit (IMU) to detect the body pose, and integrated joint-encoders and motor current sensors which are built into the servo motors. The legs use a five-bar architecture and each leg contains three actuated joints for hip abduction/adducti, hip flexion/extension, and knee flexion/extension. The end foot point can follow any path that lies inside a trapezoidal region within the planar workspace of the five-bar leg (see Fig.~\ref{fig:IK_S2}). This planar workspace is extended in 3D with the abduction motors. The calculation of the leg work-space and its inverse kinematic details can be found in \cite{dhaivatdesigndevelopment}. The URDF model used in the simulator was obtained from the SolidWorks assembly (see Fig. \ref{fig:IK_S2}). Overall, the robot model consists of $6$ floating degrees-of-freedom and $12$ actuated degrees-of-freedom.






\subsection{Walking control strategy for Stoch $2$}
\label{subsec:control_strategy}

\begin{figure}
    \centering
    \includegraphics[width=0.7\columnwidth]{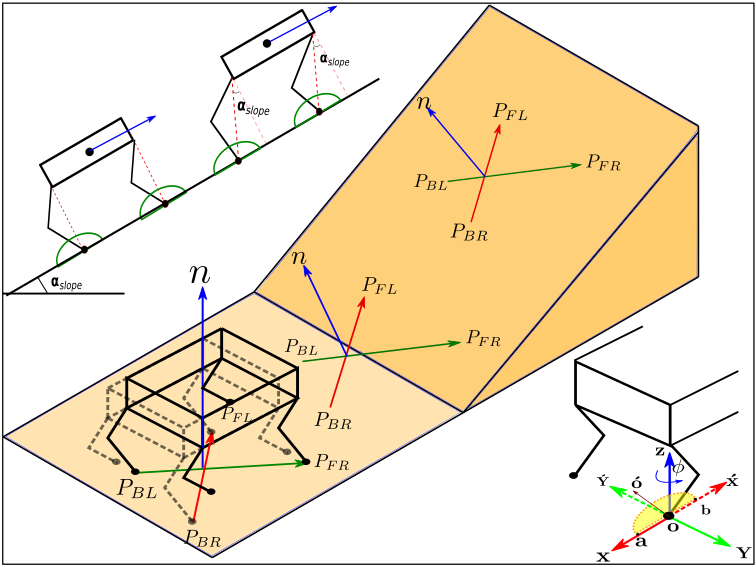}
    \caption{Figure showing the parameterization of the semi-elliptic trajectory tracked by the feet of the robot (bottom right), the foot position vectors used for the slope estimator (middle), and the two commonly used strategies for walking on slopes (top left).}
    \label{fig:semi_ellipse}
\end{figure}

As mentioned previously, we use trajectory based policies \cite{Trajectory2019Kolathaya}, i.e., we shape the semi-ellipses of every leg to realize walking in Stoch $2$. This idea of shaping of the end-foot trajectories is mainly obtained from \cite{sprowitz2018oncilla}. 
More details about the specific parameters used for shaping are provided in Fig. \ref{fig:semi_ellipse}. The general idea is to use the orientation of the robot and the supporting plane (terrain), to obtain the trajectory parameters.
We will first describe the mechanism to estimate the slope of the supporting plane, and then focus on walking controller used on arbitrary slopes.

\label{subsec:terrain_estimation}
\textbf{Terrain slope estimation.} 
Assuming a reasonably flat terrain, the surface on which the robot is walking can be modelled using a plane.  
An estimate of the terrain slope is required by the controller to enable the robot to adapt its posture  according to the terrain. This is obtained by fusing the sensor data from an Inertial Measurement Unit (IMU) and the joint encoders.
Since the stance feet make contact with the slope, a history of stance feet positions are used to determine the slope of the terrain (i.e., the slope of the plane passing through such points) w.r.t the body of the robot. The positions $\mathbf{p}_i$ of the robot feet expressed in the body frame of reference can be determined using the data from the joint encoders in the motors:
 \begin{align}
    \mathbf{p}_i = \mathbf{f}(\alpha_{hip_i}, \alpha_{knee_i}, \alpha_{abd_i}),
 \end{align}
 where $\mathbf{f}$ is the forward kinematics function that maps the joint angles to the Cartesian coordinates of the foot, $\mathbf{p}_{i}$ is the position of the foot $i$, where $i\in \{FL, FR, BL, BR\}$, in the robot's body frame of reference, and $\alpha_{abd_i}$, $\alpha_{hip_i}$ and $\alpha_{knee_i}$ are the angles of the abduction, hip and knee actuators respectively (see Fig.~\ref{fig:IK_S2}) of the leg $i$.
 
 Data from an IMU provides an estimate of the pose of the robot w.r.t. a world reference frame.
 In this case, the world reference frame is assumed to be located at the center of mass of the robot with its z-axis pointing opposite to the direction of action of gravity and with the x-axis pointing to the magnetic north. The positions of the robot feet, $\mathbf{p}_i$, can be expressed in the world reference frame, $\mathbf{P}_i$, by transforming the vectors to the new frame $\mathbf{P}_i = \mathbf{R} \mathbf{p}_i$,  where $\mathbf{R}$ is the rotation matrix that defines the orientation of the robot body frame w.r.t. the world frame. The positions of the feet should be ideally captured at the exact moment of transition when all four legs are on the ground. Due to the inability to detect the time of the transition, data is captured from the stance leg pair just before lift-off from the ground and the new stance leg pair just after touch-down. The difference between the two data capture times is small and hence the two events can be reasonably approximated to have occurred simultaneously.
%
%
The normal vector to the plane (see Fig.~\ref{fig:semi_ellipse}) can then be obtained by:
\begin{align}
    \mathbf{n} &= ( \mathbf{P}_{FR} - \mathbf{P}_{BL} ) \times 
    (\mathbf{P}_{FL} - \mathbf{P}_{BR}) .
\end{align}
The roll and pitch orientation of the plane can be readily calculated from the normal vector.

\textbf{Trajectory shaping on supporting planes.}
The strategy for walking on slopes introduces new challenges, and cannot be obtained directly from walking controllers used for horizontal planes. 
Depending on the slope, the semi-ellipses are shifted, reoriented and reshaped in real-time so that the walking is stable. The parameters for this mechanism are $\mathbf{O}\mathbf{\acute{O}}$, $\phi$, and $\mathbf{ab}$, which are shown in Fig. \ref{fig:semi_ellipse}. 
This strategy is mainly motivated from the methods presented in \cite{gehring2015dynamic}.
In particular, two types of strategies were proposed in~\cite{gehring2015dynamic}: \emph{telescopic strut} and \emph{lever}.


In telescopic strut strategy, the feet are vertically below the hip, and in lever strategy, the legs are aligned with the normal to the supporting plane (see top left of Fig.~\ref{fig:semi_ellipse}). In both of these strategies, the goal is to ensure that the torso is aligned with the supporting plane.
Owing to our learning based approach, we shape our rewards accordingly, to ensure that the torso is aligned with the support plane. While we let our controller learn the right strategy by itself, we use the telescopic strut strategy to obtain the right seeds for training (see Section \ref{subsec:policytrainingdetails} for details).
Accordingly, we model a linear policy that takes in the body and plane's orientations as the inputs and chooses the parameters of the semi-ellipse in real time, which are then tracked by the low level motor controllers.


\subsection{Reinforcement learning framework}
We formulate rough terrain locomotion as a reinforcement learning problem, with an observation space, action space and a reward function detailed below. Our policy $\pi$ is a linear transformation that maps the observation vector $s\in\R^{11}$ to the action vector $a\in\R^{20}$. 


\textbf{Observation space.}
\label{sec:obs_space}
 Our observation space is constituted of the robot base orientation and terrain slope parameters since these quantities are highly relevant observations for locomotion on slopes, as explained in Section~\ref{subsec:control_strategy}. The base orientation is measured directly by an inertial measurement unit (IMU) while the terrain slope parameters estimated by the equations provided in Section~\ref{subsec:control_strategy}. 
The observation space forms an $11$-dimensional state vector defined as $s_t$ = $\{\Theta_{t-2},\Theta_{t-1},\Theta_t,\lambda_t, \gamma_t\}$, where $\Theta_t \in \mathbb{R}^{3}$ is the base orientation (roll, pitch and yaw), $\lambda_t$ and $\gamma_t$ are the estimated roll and pitch of the slope at time step t respectively. 
It was observed that a history of body orientation at timesteps $t-2$, $t-1$ and $t$ was essential in training a robust locomotion policy (more analysis is presented in Section ~\ref{sec:results}). 
The length of the history configuration was found empirically by analyzing the final performance of the policy. 


\textbf{Action space.}
\label{subsec:actionspace}
The action space provides a set of transformations of the 2D semi-elliptic end-foot trajectories in a 3D workspace of each leg. A semi-elliptic trajectory is chosen to allow for computationally simple implementation of the controller. 
The transformations consist of translation of the semi-ellipse along $X$, $Y$, and $Z$ axes ($\mathbf{O}\mathbf{\acute{O}}$ in Fig.~\ref{fig:semi_ellipse}), rotation about $Z$-axis ($\phi$ in Fig.~\ref{fig:semi_ellipse}) and variation of the length of the major axis ($\mathbf{ab}$ in Fig.~\ref{fig:semi_ellipse}) which is also the footstep length. All the transformations are defined in the local leg frame of reference (see Fig. \ref{fig:semi_ellipse}). The action space is a $20$-dimensional vector defining these five transformations for each of the four legs. Having obtained the end-foot trajectory after the corresponding transformations, the joint angles are obtained via an inverse kinematics solver. More details about the inverse kinematics for parallel five-bar linkages are provided in \cite{dhaivatdesigndevelopment}. It is worth noting that the minor axis of the semi-ellipse is kept fixed as $0.06$m since foot clearance can be varied by moving the trajectory higher when required. The detailed description of our semi-elliptical trajectory generator can be found in the Appendix.

\section{Training and Simulation}
\label{sec:learningalgorithm}
Having defined the model and the control methodology, we are now ready to discuss the policy and training algorithm used for Stoch $2$.

\subsection{Training algorithm}

Augmented Random Search (ARS) \cite{mania2018simple} is a learning algorithm which is designed for finding linear deterministic policies and is known to be on par with other model-free RL algorithms. 
We choose the policy to be $\pi(s):=M s$, where $M\in \mathbb{R}^{20 \times 11}$ is the matrix that maps the states to actions. Let the parameters of $M$ be denoted by $\theta$. In this case, as we are not adding any constraints to the matrix $M$, the parameters $\theta$ of $M$ are simply each element of $M$. Then the goal of ARS is to determine the $\theta$, of the matrix $M$, that yields the best rewards which in turn leads to the best locomotion on non-flat terrain for Stoch 2.

ARS optimizes a parameterized policy by moving along the gradient of the function that maps the parameters of the policy to the expected reward. Since the function is stochastic in nature, different algorithms use different estimates of the gradient. ARS uses the method of finite differences as opposed to likelihood ratio methods used in other algorithms like PPO \cite{PPO} or TRPO \cite{TRPO}. We use Version $\textbf{V-1t}$ of ARS from \cite{mania2018simple}. 
We pick $N$ i.i.d. directions $\{\delta_k\}_{\{k = 1,\dots,N\}}$ from a normal distribution, where $N=20\times 11$ is the dimension of the policy parameters, i.e., $\theta \in \mathbb{R}^{N}$. 
With a scaling factor of $\nu>0$, we perturb the policy parameters $\theta$ across each of these directions. The policy is perturbed both along the positive and negative directions $\theta + \nu\delta_{(k)}$, $\theta - \nu\delta_{(k)}$. Executing this perturbed policy for one episode yields the return $R$ for each direction. Therefore, for $N$ directions, we collect $2N$ returns. We choose the best $N/2$ directions corresponding to the maximum returns. Let $\delta_{(k)}$, $k=1,2,\dots,N/2$ correspond to these best directions in decreasing order of returns. We update $\theta$ as follows:
\begin{align}
\theta := \theta +   \frac{\beta}{\frac{N}{2} * \sigma_R}\sum_{k=1}^{N/2} \left (R(\theta + \nu\delta_{(k)}) - R(\theta - \nu\delta_{(k)})\right  ) \delta_{(k)},
\end{align}
where $\beta>0$ is the step size, and $\sigma_R$ is the standard deviation of the $N/2$ returns obtained. One iteration completes when $\theta$ gets updated in the matrix $M$. In one iteration we have $2N$ episodes i.e. $440$ episodes and each episode has 400 steps.

\subsection{Reward function}
\label{sec:customreward}
We design a reward function to encourage our robot to adapt to the underlying terrain and to move in the forward direction. We have
\begin{align}
r= & \: G_{w_1}(r_{roll}- p_{roll})+G_{w_2}(r_{pitch}-p_{pitch})+G_{w_3}(r_{yaw}-d_{yaw})+G_{w_4}(h-h_d) \nonumber \\
 & + W \Delta x - S_P,
\end{align}
where $r_{\square}$ (with the square subscript being either $roll$, $pitch$ or $yaw$) is the orientation of the robot's base, $p_{\square}$ is the orientation of the support plane, $d_{yaw}$ is the desired heading, $h$ is the robot height, $h_d$ is the desired height, $\Delta x$ is the distance travelled, and $S_P$ is the standing penalty. The mapping $G: R \to [0, 1]$ is the Gaussian kernel function, and is given by
$G_{w_j}(x)=\exp{(-w_j*x^2)}, \: w_j > 0$.
Here $d_{yaw}=0$ to maintain the robot heading constant along the $X$ axis. Note that $d_{yaw}$ can also be used as an input for steering the robot in the desired direction.
We are also rewarding the robot for walking with the desired height so as to maintain sufficient clearance and avoid multiple optimal solutions for different heights. 
We are normalizing $\Delta x$ by dividing it by the maximum possible forward step length. $W$ is its corresponding weight.
With $S_P$, we are penalizing the robot for standing still, whenever the distance travelled is less than $2~$cm in $50$ time steps.

\subsection{Policy training details}
\label{subsec:policytrainingdetails}
An open-source physics engine, PyBullet, was used to simulate and train our agent. Accurate measurements of link-lengths, moments of inertia and masses were stored in the URDF file. An asynchronous version of Augmented Random Search with $15$ parallel agents was used to speed up the training on the robot. We trained our robot to walk on slopes of different inclination kept at varied orientations about the Z axis of the world frame. The inclinations are discretized  at $0^\circ,5^\circ,7^\circ,9^\circ,11^\circ$, and the orientations (yaw) in the range $0^\circ$ to $90^\circ$ degrees with $15^\circ$ resolution, making 7 different orientations for each angle of inclination except for 0 degrees, Thus giving us a grid of $29$ combinations ($7*4+1$). If the orientation is $0^\circ$ then the incline is in uphill fashion with no roll and if orientation is $90^\circ$, then the incline is in a side hill fashion with no pitch.
After every policy update, we sample an inclined slope and its orientation from the above mentioned ranges.
Instead of starting with a random policy matrix, we used a guided initial policy to give a warm start to our ARS algorithm for policy exploration.
We chose a few sets of hand-tuned fixed action values with reference to the telescopic strut strategy as discussed in Section~\ref{subsec:control_strategy}, and simulated these actions to collect corresponding observations from the environment. Then, we did supervised learning to map the observations with the hand-tuned action values to obtain our guided initial policy. With our current action framework, it is quite intuitive to come up with such sets of action values that can perform fairly better than a random initial policy which significantly reduces the number of training iterations. 

We are using the curriculum learning technique \cite{Bengio2009} for training the policy which makes the learning smooth, efficient, and effective. 
The curriculum is designed based on our intuition of tasks difficulty we begin with lower incline degrees and advances to steeper inclines. We introduce the curricula in two stages. In stage one we sampled $0^\circ$, $5^\circ$ and $7^\circ$ degree slopes with all the described orientations.
Similarly, in stage two we introduced $9^\circ$ and $11^\circ$ degree inclines after $30$ training iterations.
Also in this stage the sampling was favorable to the difficult tasks i.e steeper inclines. Similar approaches have been shown to be effective in \cite{Yu2018}, \cite{ALLSTEPS} and \cite{Park2019}.  The hyper-parameters of ARS used in training  are: Learning rate ($\beta)=0.05$, noise ($\nu)=0.04$ and episode length was of 400 steps.

\textbf{Robustness.}\label{sec:domainrand} 
We incorporate robustness by using two schemes as similar to \cite{google_paper}: 1) domain randomization and 2) applying external forces on the robot base. In domain randomization, 
we randomize the surface friction, robot’s mass distribution and the motor strength.
The friction coefficient was varied from $0.5$ to $0.8$, the additional masses of $0$ to $200~$ grams were added to the front and back portions of the robot, and motor strength was varied from $5$ to $8~$Nm in addition to a randomly chosen inclined slope as discussed in Section \ref{subsec:policytrainingdetails}. 
Similarly, we applied a random latertal force of $60-120 \mathrm{N}$ to the base of our robot for a fixed interval of 10 steps ($0.05~$s) at the mid of an episode. 

\begin{figure}%
    \centering
 \includegraphics[width=1\textwidth, height=6.5cm]{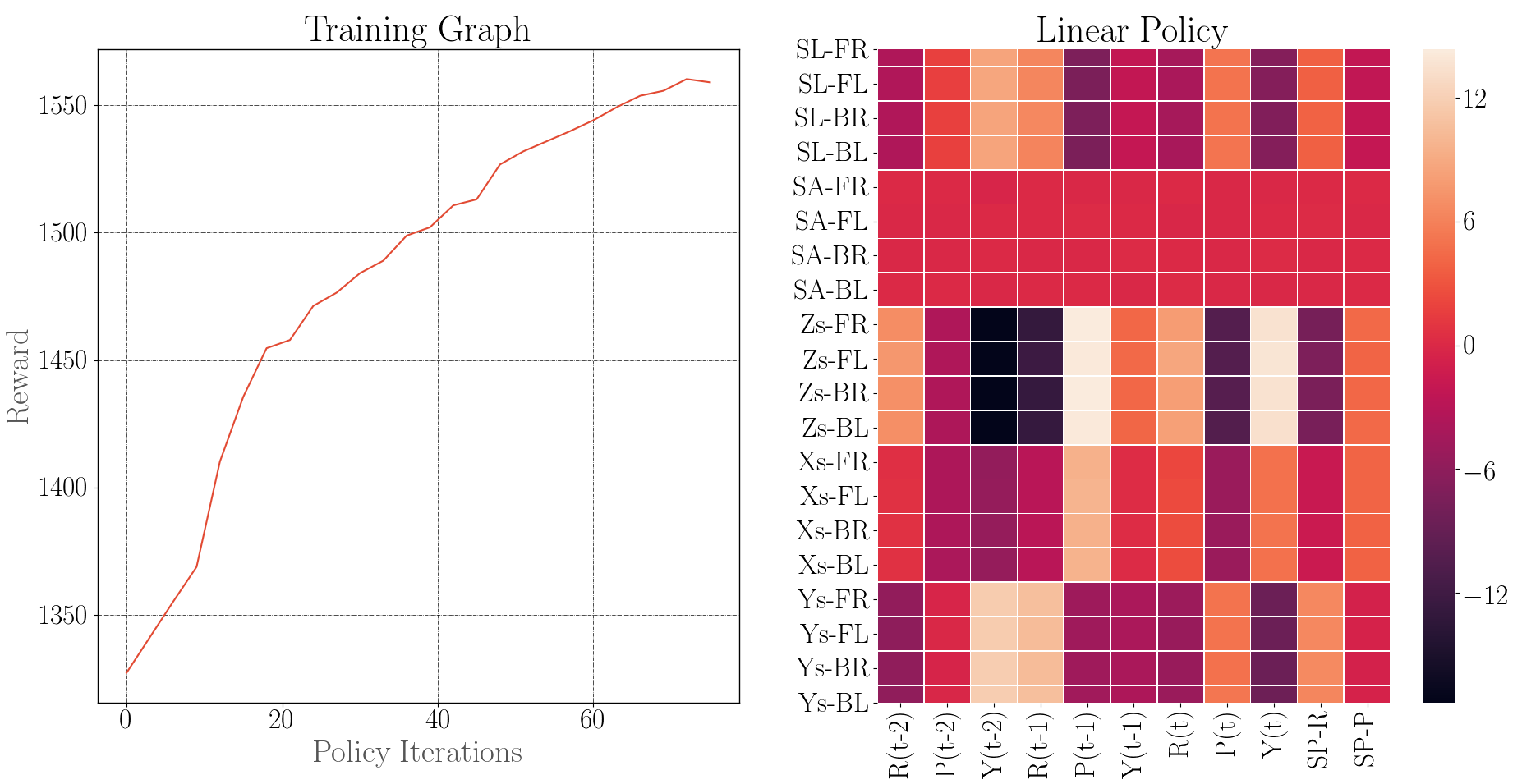}%
    \caption{Left figure shows rewards evaluated after every three policy iterations. This policy is learned in $72$ iterations. Right figure shows the heat-map of the learned matrix $M$. Here, labels $SL, SA, X_s, Y_s$ and $Z_s$ denotes the step length, steering angle, x shift, y shift and z shift respectively. $R(t), P(t), Z(t), $ denotes the Roll, Pitch, Yaw of the torso at timestep $t$. $SP$-$R$, $SP$-$P$ corresponds to  estimated support plane roll and pitch values respectively.}%
    \label{fig:rewards}%
\end{figure}

\textbf{Evaluation.} We created a small evaluation dataset that spans over the entire training dataset. We evaluated our policy after every $3$ training iterations and received the best total average reward of $1550$ in only $72$ iterations as shown in Fig. \ref{fig:rewards}. Since we began our training using a guided initial policy, we achieved higher rewards in initial iterations itself as opposed to a random initial policy.
\section{Results and discussion}
\label{sec:results}

\begin{figure}[h!]%
    \centering
      \includegraphics[trim=5 2 5 70,clip,width = 0.9\textwidth]{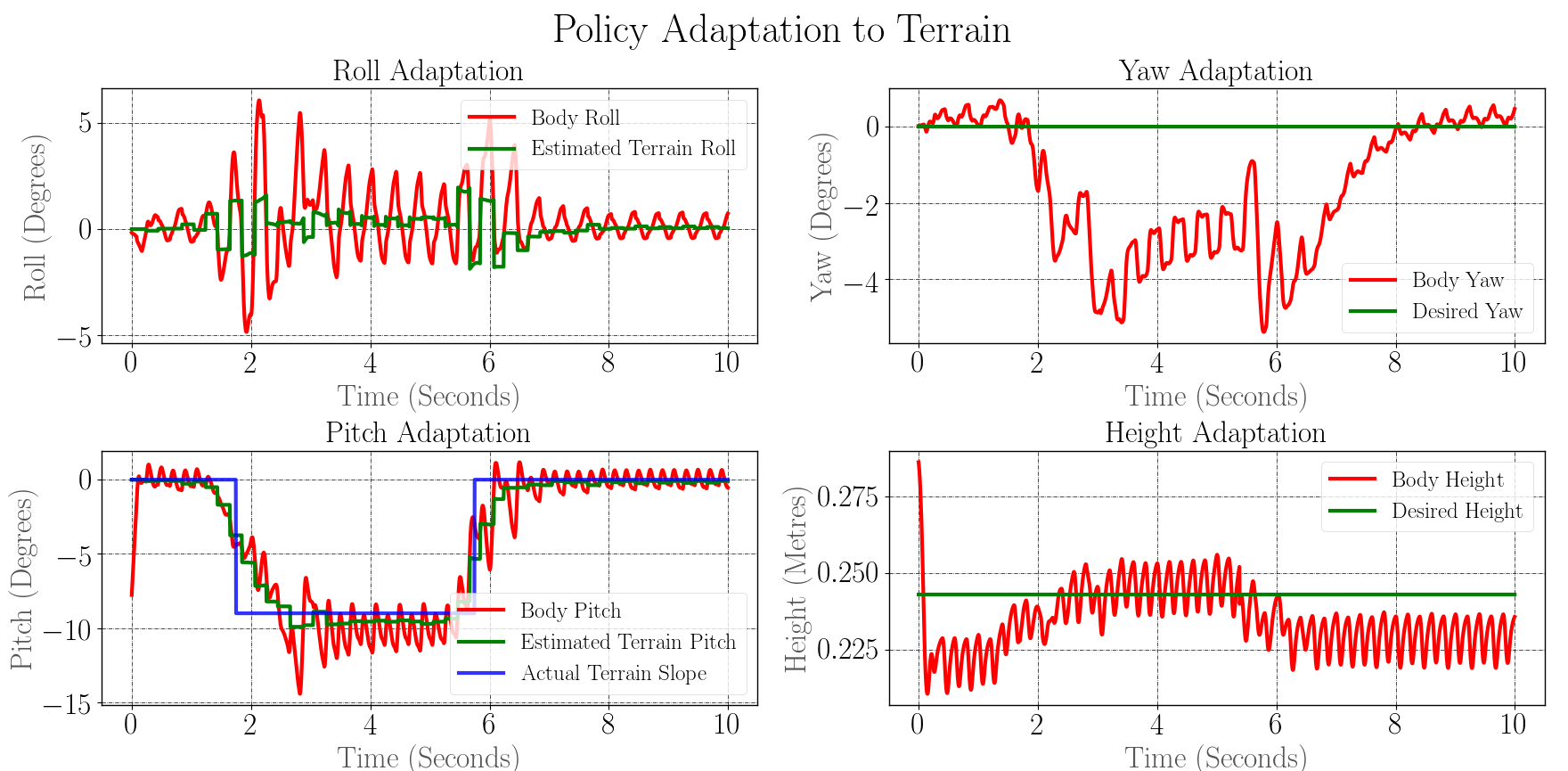}%
    \caption{Figures showing how the policy adapts the center of mass orientation and height to that of the terrain in real time. The plane is at $-9^\circ$ incline.}
    \label{fig:policy_adaptation}%
    \vspace{-3mm}
\end{figure}

With the proposed control framework, Stoch $2$ was able to perform dynamic trotting on a variety of terrain slopes (5$^{\circ}$, 7$^{\circ}$, 9$^{\circ}$, 11$^{\circ}$) kept at different orientations. Slopes  $\geq$13$^{\circ}$ are not evaluated due to the kinematic limits of our robot. 
Measurements collected during a simulation experiment are shown in Fig.~\ref{fig:policy_adaptation}, where the robot starts trotting on flat terrain and then traverses a 9$^{\circ}$ slope by moving uphill and reaches an elevated flat terrain. The estimated pitch and roll angles of the terrain, the torso orientation, as well as the height of the robot along the normal to the terrain are shown in Fig.~\ref{fig:policy_adaptation}. The robot effectively adapts its pitch angle to match the terrain, while the tracking of the roll angle is somewhat less accurate with a variation of $\pm3$ $^{\circ}$. The yaw angle (heading direction) varies between $\pm6$ $^{\circ}$ while climbing uphill but the robot is able to correct itself later.
 
\begin{figure}[h!]%
    \centering
 \includegraphics[trim=5 5 5 60, clip,width = 0.9\textwidth]{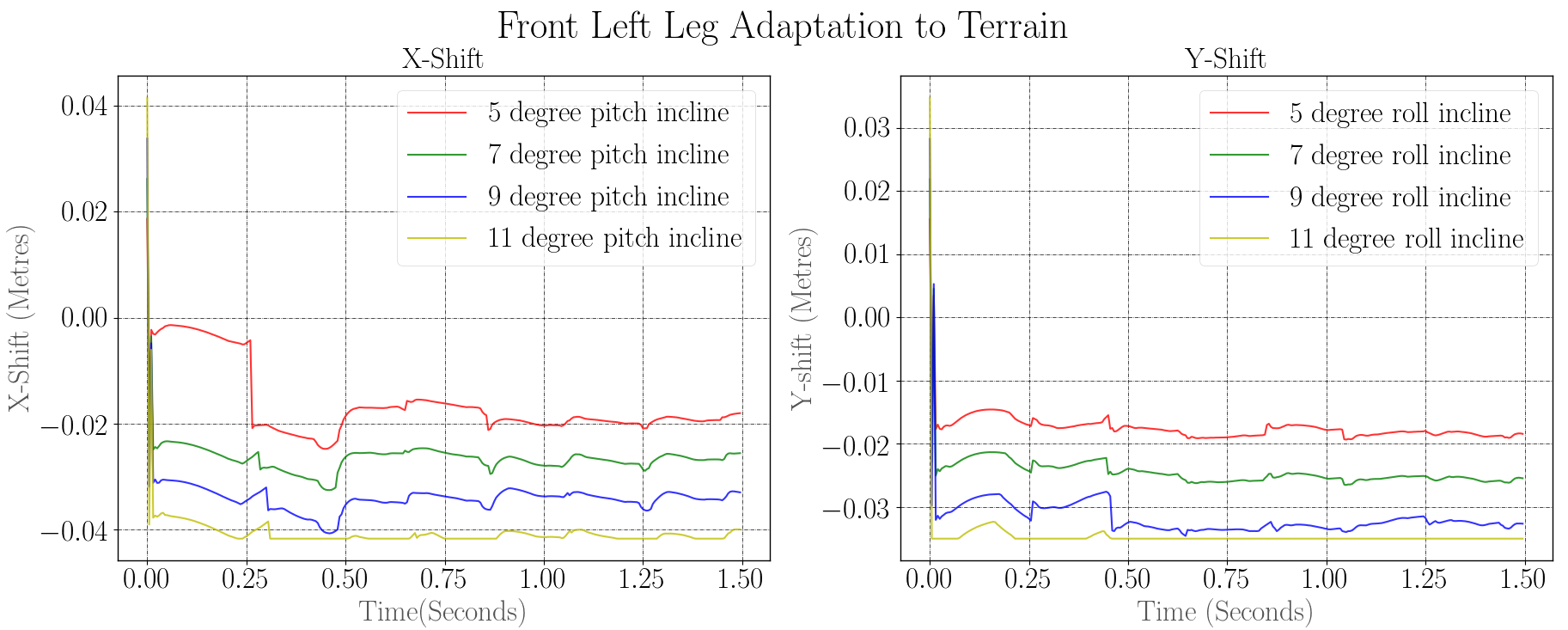}%
    \caption{This figure shows how the response of the front left leg varies with different inclines. The left figure is for an up-down incline and shows the X-Shift parameter of the trajectory while the right figure is for a left-right incline and shows the Y-Shift parameter of the trajectory.}%
    \label{fig:XandYshift}%
    \vspace{-3mm}
\end{figure}

 Upon analysing the actions in Fig.~\ref{fig:XandYshift} shows that the evolved policy shows similar characteristics as that of the \textit{telescopic strut} strategy. In Fig.~\ref{fig:XandYshift}, the left figure shows the $X$ shifts in the trajectory of the front left foot when the robot walks uphill. The $X$ shift is increasing in a negative direction with the steepness in slopes. The figure on the right shows a similar pattern in $Y$ shifts when the robot walks sidehill. The abduction joint limit was reached for a sideways slope of 11 degrees. This corresponds to the maximum allowed lateral shift (Y-shift) of 3.5 cm owing to the robot's kinematic limits. The linear policy (see Fig. \ref{fig:rewards}) is capable of generalizing to negative roll and positive pitch environments as well. This is demonstrated in the supplementary video.

  

\begin{figure}[h!]%
    \centering
 \includegraphics[trim=5 5 5 58, clip, width = 0.9\textwidth]{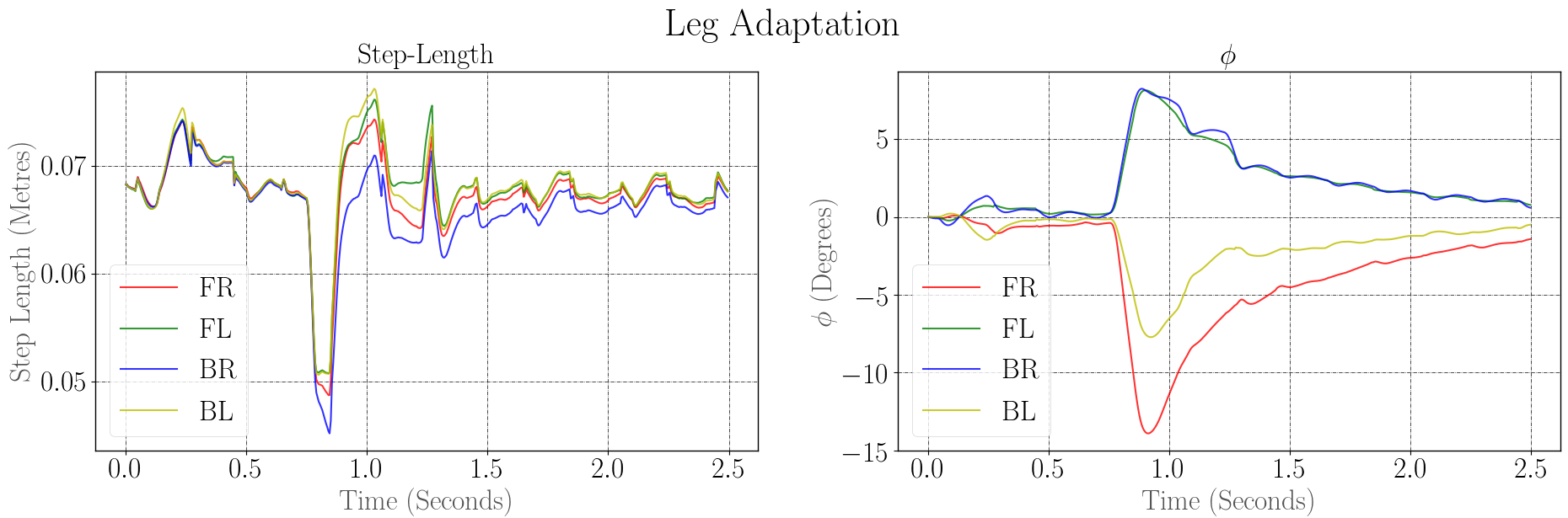}%
    \caption{This figure shows how the legs adapt to external disturbances. The left figure indicates the step-length while the right figure indicates the steering angle. This is for a $-9^\circ$ incline.}%
    \label{fig:Legadapt}%
    \vspace{-4mm}
\end{figure}

We demonstrate robustness to external disturbances by applying forces externally. The results of this are shown in Fig.~\ref{fig:Legadapt}. From time $0.7$ to $0.9$ second, a lateral external force of $100$N is applied onto the robot. This results in a change in the body orientation, which is then read by the linear policy to accordingly correct its direction of walking. As shown in Fig. \ref{fig:Legadapt}, the step length and steering quickly adjust themselves within $0.5~$s. 
In the video, we demonstrate robustness to external forces on arbitrary inclines. We also demonstrate robustness to changes in friction, masses and motor strengths on the robot, thereby showing that domain randomization has made the linear policy more robust.

Our experiments show that orientation history was essential in training a locomotion policy because it increases the representational capacity of a linear policy, allowing it to learn more complex behaviours as explained in \cite{rajeswaran2017towards}. Secondly, it enables the policy to detect foot slip and filter the sensor noise.

\textbf{Other robots.}
We verified our approach on other robots with different physical properties as compared to Stoch $2$. Owing to the simplicity of our control framework, the mere scaling of the action space values was sufficient to match the kinematic limits of the new robot. The PD gains and the frequency of the trajectory controller should also be updated accordingly. It is worth noting that no additional manual tuning is required to train a new policy. We validated our controller in robots namely, HyQ \cite{Semini2011} and Unitree-Laikago that belong to the large and medium form factors respectively. The trained policies were found to generalize well on multiple inclinations ($0^\circ-15^\circ$) and across varied orientations ($0^\circ-45^\circ$) as demonstrated in the supplementary video.
  
\textbf{Conclusion.}
We successfully demonstrated a single linear policy for quadrupedal locomotion on multiple types of sloped terrains, which is learnt via Augmented Random Search (ARS).
By assuming stairs to be sloped terrains as done in \cite{Bledt2018MIT}, our policy is capable of traversing stairs as shown in the attached video. 
The linear policy takes the body and plane orientation as inputs and provides the trajectory parameters as outputs.
This type of policy is easy to deploy on hardware, and, at the same time, do not require large computational resources, unlike the Deep Neural Network (DNN) based control algorithms existing in the literature. Our approach can be used to quickly train linear policies for multiple robots, thus significantly simplifying the process of controller design and implementation for rough terrain walking on quadruped robots. Future work will involve testing the controller in hardware. The video submission accompanying this paper is shown here: \href{https://youtu.be/KdQn1e3rI7o}{youtu.be/KdQn1e3rI7o}, and the code is released here: \href{https://github.com/StochLab/SlopedTerrainLinearPolicy}{github.com/StochLab/SlopedTerrainLinearPolicy}.









\bibliography{references}
\section{Appendix}

In a trot gait the diagonally opposite legs move in phase while the other two move out of phase. This means that the swing phase and stance phase of the legs last for the same duration. If we assume that the stance phase (the flat portion of the semi-elliptic trajectory) for any leg lasts for the first half of its cycle and the swing phase for the next half cycle, then taking the hip as the origin, we have the $x,y,z$ position of the foot as:
\begin{align}
    \text{stance} &: x = 0.5*l\cos{(2\pi(1-\tau))}, \:\:
    &y = 0, \:\:
    &z = -h_d, \quad &\tau \in [0, 0.5), \nonumber \\
    \text{swing} &: x = 0.5*l\cos{(2\pi(1-\tau))}, \:\:
    &y = 0, \:\:
    &z = -h_d + f_c\sin{(2\pi(1-\tau))}, \quad &\tau \in [0.5, 1). \nonumber
\end{align}
where $l$ is the step-length ($=\mathbf{ab}$ from Fig. \ref{fig:semi_ellipse}), $h_d$ is the desired height along the support plane normal, $f_c$ is the foot clearance and $\tau$ is the time normalized w.r.t. time period $T$ of one cycle. For Stoch 2, $l \leq 0.136m$, $h_d = 0.243m$, $f_c = 0.06m$ and $T = 0.4s$. The semi-elliptical trajectory which lies in the $X$-$Z$ plane is then transformed according to the action by:
\begin{align}
    x^\prime = x_s + x \cos(\phi), \quad
    y^\prime = y_s + x \sin(\phi), \quad
    z^\prime = z_s + z,
\end{align}
where $x^\prime$, $y^\prime$ and $z^\prime$ are the coordinates of the foot in the leg frame of reference after the transformation, $x_s$, $y_s$ and $z_s$ are the shift (given by $\mathbf{O\acute{O}}$ in Fig. \ref{fig:semi_ellipse}) in the three corresponding Cartesian directions as obtained from the controller, and $\phi$ is the yaw rotation of the semi-ellipse (see Fig. \ref{fig:semi_ellipse}). $x',y',z'$ are then represented in the common body frame by appropriate transformations.

\end{document}